\pdfoutput=1

\documentclass[10pt,twocolumn]{article}
\usepackage[a4paper,margin=2.0cm]{geometry}

\usepackage[utf8]{inputenc} 
\usepackage[T1]{fontenc}
\usepackage{lmodern}

\usepackage{graphicx}
\usepackage{booktabs} 
\usepackage{amsmath}
\usepackage{array}
\newcommand{\PreserveBackslash}[1]{\let\temp=\\#1\let\\=\temp}
\newcolumntype{C}[1]{>{\PreserveBackslash\centering}p{#1}}
\newcolumntype{R}[1]{>{\PreserveBackslash\raggedleft}p{#1}}
\newcolumntype{L}[1]{>{\PreserveBackslash\raggedright}p{#1}}
\newcommand{\drow}[1]{\multirow{2}{*}{#1}}
\usepackage{multirow}
\usepackage[hyphens]{url}
\usepackage{hyperref}
\hypersetup{
    colorlinks=true,linkcolor=black,citecolor=black,urlcolor=blue,
    pdfauthor={Wagner Gonçalves Pinto},
    pdftitle={On the reproducibility of fully convolutional neural networks for modeling time-space evolving physical systems}
}
\usepackage{textcomp}

\usepackage{caption}
\title{
    On the reproducibility of fully convolutional neural networks for modeling time-space evolving physical systems
}

\date{2021}

\author{
    Wagner Gonçalves Pinto\thanks{Aerodynamics, Energetics and Propulsion Department, ISAE-SUPAERO,
Université de Toulouse, Occitanie, Toulouse, France}
    \\
    \and 
    Antonio Alguacil\footnotemark[1]
    \thanks{Mechanical Engineering Department, Université de Sherbrooke, Sherbrooke, Quebéc, Canada}
    \\
    \and 
    Michaël Bauerheim\footnotemark[1]
    \\
}

\begin{document}

\maketitle

\begin{abstract}

Reproducibility of a deep-learning fully convolutional neural network is evaluated by training several times the same network on identical conditions (database, hyperparameters, hardware) with non-deterministic Graphics Processings Unit (GPU) operations. The propagation of two-dimensional acoustic waves, typical of time-space evolving physical systems, is studied on both recursive and non-recursive tasks. Significant changes in models properties (weights, featured fields) are observed. When tested on various propagation benchmarks, these models systematically returned estimations with a high level of deviation, especially for the recurrent analysis which strongly amplifies variability due to the non-determinism. Trainings performed with double floating-point precision provide slightly better estimations and a significant reduction of the variability of both the network parameters and its testing error range.

\end{abstract}

\section{Motivation and context}
\label{sec:context}

Reproducibility, the principle that a method or experiment can be replicated in different conditions, is essential in science. Obtaining reproducible results is complex in deep learning (DL) due to the large databases, the intrinsic stochastic nature of the majority of the algorithms, the complex nature of the optimization search-spaces and the use of non-deterministic computations. As such, this topic is constantly being discussed~\cite{Peng2011,Ivie2018,Renard2020,Pineau2020}.

Regarding the computational source of variability, numerical non-determinism comes from the rounding of numbers associated with a stochastic order of arithmetic operations~\cite{Whitehead2011,Chou2020}. Its is notably observed when the same code is evaluated in different hardware~\cite{Jezequel2015} or with different floating point precision~\cite{Seznec2018}. Currently, performance requirements make the use of Graphics Processing Units (GPUs) mandatory in DL, their major drawback being the difficulty to perform deterministic operations. Different runs in the same GPU can also be non-deterministic due to use of algorithms that favor performance over repeatability~\cite{Iakymchuk2015,Jorda2019}.

The use of neural networks for estimating Partial Differential Equations (PDEs) is a growing trend and has been widely evaluated in the Fluid Mechanics field, many examples are available on the review by Brunton et al.~\cite{Brunton2020}; general purpose PDE solvers are also discussed on the literature~\cite{Berg2018,Sirignano2018}. In contrast with classification problems, the estimation of spatial-temporal PDEs must guarantee pixel-wise precision. Even if determinism is not mandatory, levels of accuracy and reproducibility must be mastered, and thus, further evaluated and discussed.

This paper presents a detailed comparison between several DL models that estimate the evolution of acoustic waves, as a benchmark for time-space evolving systems, trained in the same context (identical database, hardware, random number seeds, hyperparameters, software and hardware) in a GPU with non-deterministic operations. Similar analysis in the context of Reinforced Learning~\cite{Nagarajan2018} found that the GPU non-determinism has an impact analogous to changing the starting model’s parameters. Tests are performed for evaluating and comparing the generalization capacity of the trained models. To the authors knowledge, no previous work has discussed the reproducibility of a neural network when estimating time-space evolving systems.

Document is organized as follows. Methodology is presented in Section~\ref{sec:methodology}, with the description of the physical system, the network architecture, and the training and testing procedures. Models properties (losses convergence, kernels weights and featured fields) are discussed in Section~\ref{sec:models}. In Section~\ref{sec:recurrent}, results for the recurrent analysis are presented for solution benchmarks and a database with random acoustic pulses. Section~\ref{sec:conclusions} closes the document with final remarks. Code, descriptions of the neural network and the numerical setup and examples of the estimated fields are available in the supplementary material\footnote{Supplementary material is available at: \href{https://github.com/wagnifico/cnn-phys-reproducibility}{github.com/wagnifico/cnn-phys-reproducibility}}.

\section{Methodology}
\label{sec:methodology}

\subsection{Physical system and solver}
\label{sec:simulation}

This work focuses on the propagation of two-dimensional (2D) acoustic waves in homogeneous medium. This is a relatively simple problem to be used as a benchmark of a typical time-space evolving physical system. More complex setups can be derived from this framework (three dimensional, presence of obstacles, complex boundaries), where computational costs with classical solvers start to be significant. Neural networks are believed to provide representative gains in performance while maintaining similar accuracy.

Acoustic waves are simulated using the Lattice Boltzmann Method (LBM). Macroscopic quantities such as density and velocity are obtained by modeling the microscopic distribution of particles described by the Boltzmann equation, more details on the method can be found in~\cite{Kruger2016}. Simulations are performed using the open source Palabos\footnote{Palabos solver: \href{https://palabos.unige.ch/}{https://palabos.unige.ch/}} framework~\cite{Palabos2020}. Density fields are initiated with Gaussian pulses:

\begin{small}
    \begin{equation}
        \rho(\mathbf{x},t=0) = \left\{
            1 + \varepsilon~\mathrm{exp}\left[ 
                - \mathrm{log}(2)~d(\mathbf{x},\mathbf{x}_c)^2/h_w^2
                \right]
            \right\}
            \rho_0
        \label{eq:LBM_gaussian_pulse}
    \end{equation}
\end{small}

\noindent where $\rho$ is the density, ${\mathbf{x} = (x,y)}$ are the space coordinates, $t$ is the time, $d$ is the Euclidean distance from the center of the pulse $\mathbf{x}_c$, $\varepsilon$ is the pulse initial amplitude and $h_w$ is the half-width at half maximum of the Gaussian. Current analysis considers pulses of fixed amplitude and size (0.001 and ${12 \Delta x}$ , respectively). The acoustic density $\rho\rq$ is considered, where ${\rho\rq = \rho - \rho_0}$, being the ambient density set as ${\rho_0 = 1.0}$, LBM units. Domain is square, discretized by ${N = 200}$ uniform grid points in both directions, with imposed null velocity at boundaries (reflecting). Each timestep represents a duration of ${\Delta t = 0.0029~\ell/a}$, where ${\ell = 100~\mathrm{m}}$ is the domain size and ${a=343 ~\mathrm{m/s}}$ is the sound speed. This means that 173 timesteps, i.e., 43 recurrences (a recurrence being a jump of 4 timesteps, as explained in Section~\ref{sec:NN}) are necessary for a wave to reach the boundaries from the center of the domain.

\subsection{Neural network architecture}
\label{sec:NN}

As most time-evolving physical problems, acoustic propagation does not require a lot of time-memory. Only a few snapshots need to be remembered in order to reconstruct approximation of the partial derivatives included in the PDEs. In that context, RNN (Recurrent Neural Network) and LSTM (Long Short-Term Memory) have been proved less effective~\cite{Fotiadis2020} for this recursive task compared with a CNN approach with a few previous snapshots as input. This latter strategy is therefore used here to output the next time frame.

The multi-scale convolutional neural network introduced by Mathieu et al.~\cite{Mathieu2016} is used. As indicated by the latter authors, the major limitation of a CNN is that it only accounts for short-range dependency, limited to the size of the convolution kernels. The use of multi-scale network is capable of surpassing this limitation by considering a different structure size inside each scale. Current network is composed by 3 scales ($N$, ${N/2}$ and ${N/4}$) and the transitions between them (upsampling and downsampling) are performed using bi-linear interpolations. Each scale follows the principle of the Laplacian pyramid used in image compression~\cite{Denton2015}, similar to an U-Net~\cite{Ronneberger2015}.

\begin{figure}[t!]
    \begin{center}
        \includegraphics
            {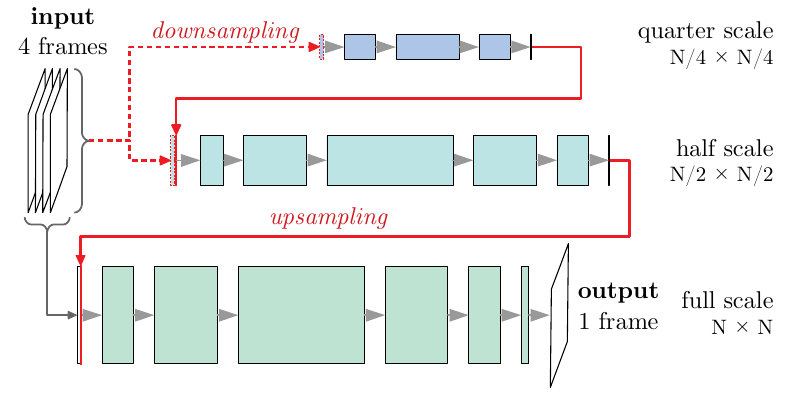}
        \caption{
            Simplified diagram of the multi-scale neural network. Arrows between boxes indicate a two-dimensional convolution operation, boxes' height and breadth are proportional to the frame dimension and number of layers, respectively; interpolations operations are presented by dashed lines for downsampling and continuous lines for upsampling.
            }
        \label{fig:diagram_NN}
    \end{center}
\end{figure}

Replication padding is used at the borders in order to have a constant layer size. Non-linearity is added by applying the $\mathrm{ReLU}$ activation function. Due to the nature of the problem, they are not present for the totality of layers so that negative density values can be obtained. A simplified representation of the neural network is presented in Figure~\ref{fig:diagram_NN}. It is composed by a total of 422,419 trainable parameters, divided in 17 convolutions. Model is coded using the PyTorch\footnote{PyTorch framework: \href{https://pytorch.org/}{https://pytorch.org/}} framework~\cite{pyTorch}.

\begin{figure}[t!]
    \begin{center}
        \includegraphics
            {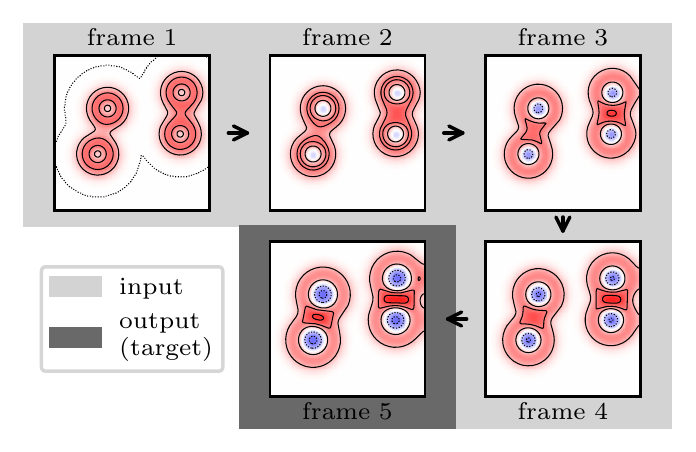}
        \caption{
            Sample of a datapoint, composed by 5 scalar fields of acoustic density in LBM units,
            4 as input and 1 as output (target). Iso-lines are selected for
            intervals of ${\pm 0.125~\varepsilon}$, starting at 0 (continuous for
            positive, dotted for negative, zero omitted); arrows indicate the
            chronological order.
            }
        \label{fig:field_sample_datapoint}
    \end{center}
\end{figure}

Input is composed by 4 acoustic density fields, called frames, issued from the LBM simulations, ordered in time. Output is the following frame, representing the field to be estimated. A jump of 4 simulation timesteps is considered between each frame. A datapoint is a group of these 5 frames, as illustrated in Figure~\ref{fig:field_sample_datapoint}. The simulated fields are split with no overlap, that is, every datapoint has a group of unique frames. At each forward pass, data of all frames is divided by the standard deviation of the first frame before performing the calculations. Loss, presented next, is equally scaled.

As in \cite{Mathieu2016}, the L2-norm of the scalar fields and their spatial gradients compose the loss $\mathcal{L}$:

\begin{small}
    \begin{equation}
        \begin{gathered}
            \mathcal{L} = \frac{1}{M}
                \sum_{k=1}^{M}
                \left(
                    \lambda_\mathrm{L2} \mathcal{L}_\mathrm{L2}
                        + \lambda_\mathrm{GDL} \mathcal{L}_\mathrm{GDL}
                \right)
            \\
            \mathrm{where}
            \\
            \mathcal{L}_\mathrm{L2} = (\rho\rq_k - \tilde{\rho}\rq_k)^2
            \\
            \mathcal{L}_\mathrm{GDL} =
                \left[
                    \frac{\partial}{\partial x}
                        \left(
                            \rho\rq_k - \tilde{\rho}\rq_k
                        \right)
                    \right]^2
                +
                \left[
                    \frac{\partial}{\partial y}
                        \left(
                            \rho\rq_k - \tilde{\rho}\rq_k
                        \right)
                    \right]^2
        \end{gathered}
        \label{eq:NN_loss}
    \end{equation}
\end{small}

\noindent where $\tilde{\rho}\rq$ is the estimated acoustic density, subscript $k$ represents a grid point (or pixel), ${M = N \times N}$ is the number of grid points and $\lambda$ are the loss factors, set as ${\lambda_{L2} = 0.98}$ and ${\lambda_{\mathrm{GDL}} = 0.02}$. Space derivatives are calculated using a central finite difference $\mathrm{4^{th}}$ order scheme, degenerated to $\mathrm{2^{nd}}$ order (forward and backward) close to the boundaries. There are no constraints or losses associated with the hidden layers.

Weights are initialized randomly following a He initialization~\cite{He2015}. Adam optimizer~\cite{Kingma2015} is used, with a starting learning rate of $10^{-4}$. The learning rate is scheduled to reduce by 2\% after 10 epochs without loss modification.

\subsection{Training procedure and databases}
\label{sec:training}

The density fields are initiated with 1 to 4 Gaussian pulses, as described by Equation~\ref{eq:LBM_gaussian_pulse}, centered inside ${[0.1\ell,~0.9\ell]}$, both directions. Both the number and position of the pulses are defined randomly following uniform distributions.

The training and validation databases are build by 500 simulations (400 for training and 100 for validation), with a total of 4400 datapoints (3200 + 1200); combined databases size is 6.5 GB. During the learning phase, batches of 32 datapoints (160 frames) are considered. Random rotation operations by multiples of $\pi/2$ are performed for the group of fields composing the datapoints before the training. That is, there is no virtual expansion of the database, only more diversity is added to it.

Reproducibility of the network is evaluated by performing several trainings with the exact same database and hyper-parameters. Seeds of random number generators are equally set at the beginning of each training. For current framework, deterministic algorithms are not available for all the operations present in the model~\cite{pyTorch_determinisitic_doc}. Even if determinism could be achieved by using another framework or model architecture, the use of non-deterministic algorithms is deliberate as it mimics training the same model on different devices or distinct versions of libraries and drivers, where determinism may not be guaranteed~\cite{cuda_CUDNN_doc, cuda_cuBLAS_doc}.

Since non-determinism is associated with truncation error, training is done considering two floating-point precision:  single (32-bits, framework’s default, referrenced as $\mathrm{FP32}$) and double (64-bits, $\mathrm{FP64}$). Model and database are exactly the same, framework default type is modified at the beginning of the runs. Data is generated in double precision and converted to single precision for the single precision trainings. The number of runs (a run is defined as an independent training) is of 10 for single precision and 5 for double. Such number of experiments is in accordance with number of runs used in traditional hyperparameter optimization~\cite{Li2020}.

A total of 1500 epochs are calculated on each run. In order to have a representative range of the observed behaviors, models are saved at every 125 epochs so a total of 12 checkpoints are available at the time of the post-processing. Training is performed on a NVIDIA V100 GPU. The average time for the completion of an epoch (training + validation) is of about 30 seconds in single precision and 100 seconds in double precision, so respectively about 12 and 40 hours are necessary per run.

\subsection{Testing procedure and databases}
\label{sec:testing}

\begin{figure}[t!]
    \begin{center}
        \includegraphics[trim=0.0cm 0.25cm 0.0cm 0.25cm,clip]
            {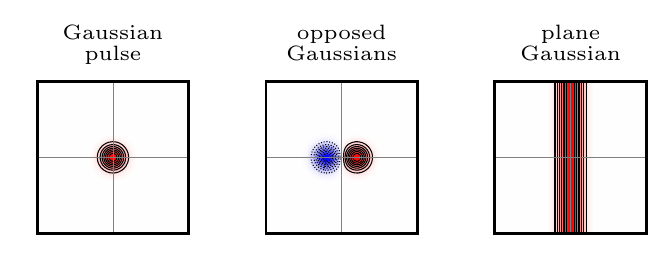} 
        \caption{
            Acoustic density fields for the benchmarks at the start of the simulation. Iso-lines are selected for intervals of ${\pm 0.125~\varepsilon}$, starting at 0 (continuous for positive, dotted for negative, zero omitted); thin gray lines indicate the domain’s symmetry axis.
            }
        \label{fig:fields_benchmarks_start}
    \end{center}
\end{figure}

Two distinct testing datasets are used for the evaluation of the quality of the trained models:

\textbf{Benchmarks database:} Three simulations with benchmarks that are unknown to the network, aiming at evaluating the capacity of generalization of the models. The benchmarks are illustrated in Figure~\ref{fig:fields_benchmarks_start} and described next (pulse half-width of ${h_w = 12 \Delta x}$ is maintained for all cases):

\begin{itemize}
    \item single Gaussian pulse at the center of the domain, amplitude of ${\varepsilon = 0.001}$ (same properties of the Gaussian pulses used for generating the trainining database);

    \item two opposite Gaussian pulses (${\varepsilon = \pm 0.001}$), placed at the $y$ symmetry axis with ${\pm 20 \Delta x}$ of offset from the center of the domain, since only positive pulses have been seen during training; and 

    \item a plane wave of Gaussian contour in $x$ (${\varepsilon = 0.001}$), extruded on $y$, which is known to be critical when evaluating the generalization capabilities of neural networks on such a problem~\cite{Sorteberg2018,Fotiadis2020}.
\end{itemize}

\textbf{Random database:} A set of 100 simulations with random Gaussian pulses, following the procedure used for generating the training and validation datasets (see Section~\ref{sec:training}). This database allows the production of statistics regarding the quality and variability of the models.

\subsection{Non-recurrent and recurrent analysis}
\label{sec:recurrent_analysis}

The network only accounts for one future frame. Estimation of further wave propagation is done via an recurrent (auto-regressive) analysis, where output of a given model evaluation is used as input in following estimations, as performed in previous works in the literature~\cite{Lee2019,Lee2019a,Alguacil2020}:

\begin{small}
    \begin{equation}
        \tilde{\rho}\rq^{f} = 
        \mathcal{F} \left(
                \tilde{\rho}\rq^{f-1},~
                \tilde{\rho}\rq^{f-2},~
                \tilde{\rho}\rq^{f-3},~
                \tilde{\rho}\rq^{f-4}
            \right)
        \label{eq:NN_recurrent}
    \end{equation}
\end{small}

\noindent where $\mathcal{F}$ represents the neural network operator and the exponents the frame number. Preliminary analysis has shown that there is an average shift of the fields when performing the recurrent analysis. This effect is contained by a posterior physics based energy preserving correction, as in~\cite{Alguacil2020}.

\section{Analysis of models}
\label{sec:models}

\subsection{Convergence of trainings}
\label{sec:convergences}

\begin{figure}[b!]
    \begin{center}

        \begin{tabular}{ C{0.8cm} C{3.0cm} C{3.00cm} }
            & {\small {\sc single ($\mathrm{FP32}$) } } & {\small {\sc double ($\mathrm{FP64}$)} } \\
        \end{tabular}
        \vskip -0.1in

        \begin{tabular}{ C{4.0cm} L{3.0cm} }
                \includegraphics[trim=0.3cm 0.0cm 0 0.0cm,clip]
                    {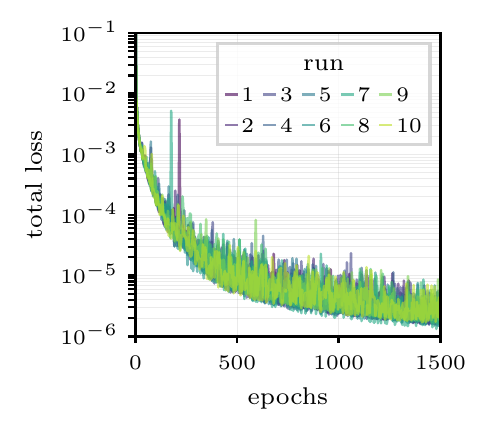}
             &
            
                \includegraphics[trim=1.25cm 0 0 0.0cm,clip]
                    {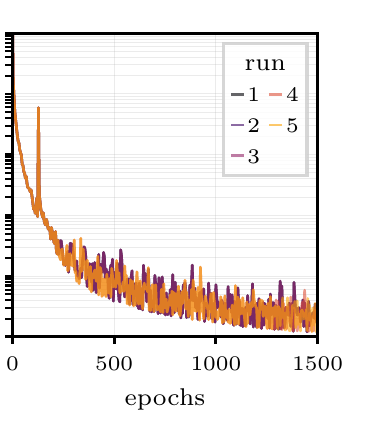}
             \\

        \end{tabular}
        \caption{
            Evolution of the training losses for the multiple runs in single (top) and double (bottom) precision. For visualization purposes, only one in every 4 values are shown on the graphs, so that the global trend is presented while allowing to see the small deviations associated with each run.
            }
        \label{fig:graph_loss_convergence}
    \end{center}
\end{figure}

The convergence of the training loss for the two groups of runs (single and double precision) is presented in Figure~\ref{fig:graph_loss_convergence}. Trainings share a similar overall level of convergence and slope; the peak around 100-200 epochs is due to a reduction of the learning rate.

For the case of single precision, the losses differ after a small number of epochs and 10 curves are displayed in the graph. For the evaluations with double precision, losses evolve identically until around 250 epochs. After that, only 3 curves can be seen (runs 1 to 3 returned identical losses). This result is a first insight on how floating precision affects the training reproducibility.

Best models for each run, considering the lowest training and validation losses, are listed on Tables~\ref{tab:models_single} and \ref{tab:models_double}. Variation is significant for single precision trainings, while the double precision runs are almost identical. Standard deviation of the losses is about 5 times smaller for the latter group.

Differences between the models are further discussed by comparing the weights and featured fields for a sample input. On these analysis and in the performance study (Section~\ref{sec:recurrent}), the best models for each run (state associated with the lowest validation loss) are considered.

\begin{table*}[t]
    \caption{Training and validation total losses of the best models for each run of the single precision ($\mathrm{FP32}$) trainings.}
    \label{tab:models_single}
    \begin{center}
        \begin{small}
            \begin{sc}
                \begin{tabular}{L{1.40cm} *{10}{C{0.75cm}} *{3}{C{0.825cm}}}
                    \toprule                
                    & \multicolumn{10}{c}{total loss, {\tiny $\times 10^{-6}$} {\tiny(epoch)}} & \multicolumn{3}{c}{statistics} \\
                    \midrule
                    \drow{run} 
                          & \drow{1} & \drow{2} & \drow{3} & \drow{4} & \drow{5} & \drow{6} & \drow{7} & \drow{8} & \drow{9} & \drow{10} & avg, & std, & \drow{ $\mathrm{\frac{MAX}{MIN}}$ } \\
                          & & & & & & & & & & 
                          & {\tiny $\times 10^{-6}$} & {\tiny $\times 10^{-7}$} & \\

                    \midrule
                    \drow{training} 
                        & 1.593  & 1.559  & 2.548  & 2.615  & 1.797  & 1.460  & 1.585  & 1.666  & 2.067  & 3.217
                            & \drow{2.011} & \drow{5.605} & \drow{2.204} \\
                        & {\tiny(1499)} & {\tiny(1499)} & {\tiny(1249)} & {\tiny(1499)} & {\tiny(1374)} & {\tiny(1499)} & {\tiny(1249)} & {\tiny(1499)} & {\tiny(1249)} & {\tiny(874)} 
                            &  &  & \\
                    \drow{validation}
                        & 2.890  & 2.792  & 3.104  & 3.030  & 3.263  & 2.481  & 2.361  & 2.880  & 3.522  & 2.868 
                            & \drow{2.919} & \drow{3.245} & \drow{1.491}  \\
                        & {\tiny(1499)} & {\tiny(1499)} & {\tiny(1249)} & {\tiny(1374)} & {\tiny(1374)} & {\tiny(1499)} & {\tiny(1499)} & {\tiny(1499)} & {\tiny(1249)} & {\tiny(1499)}
                            & & & \\
                    
                    \bottomrule
                \end{tabular}
            \end{sc}
        \end{small}
    \end{center}
    \vskip -0.2in
\end{table*}

\begin{table}[t!]
    \caption{Training and validation total losses of the best models for each run of the double precision ($\mathrm{FP64}$) trainings.}
    \label{tab:models_double}
    \begin{center}
        \begin{small}
            \begin{sc}
                \begin{tabular}{ L{1.40cm} *{5}{C{0.8cm}} }
                    \toprule                
                    & \multicolumn{5}{c}{total loss, {\tiny $\times 10^{-6}$} {\tiny(epoch)}} \\
                    \midrule
                    run & 1 & 2 & 3 & 4 & 5 \\
                    \midrule
                    \drow{training} 
                        & 1.198  & 1.198  & 1.198  & 1.312  & 1.237 \\
                        & {\tiny(1499)} & {\tiny(1499)} & {\tiny(1499)} & {\tiny(1499)} & {\tiny(1499)} \\
                    \drow{validation}
                        & 2.061  & 2.061  & 2.061  & 2.057  & 2.077 \\
                        & {\tiny(1499)} & {\tiny(1499)} & {\tiny(1499)} & {\tiny(1499)} & {\tiny(1499)} \\                    
                \end{tabular}
                \begin{tabular}{ L{1.40cm} *{3}{C{1.60cm}} }
                    \midrule               
                    & \multicolumn{3}{c}{statistics} \\
                    \midrule                    
                    & avg, {\tiny $\times 10^{-6}$} & std, {\tiny $\times 10^{-7}$}& max/min \\
                    \midrule
                    training & 1.229 & 0.446 & 1.096 \\
                    validation & 2.063 & 0.071 & 1.010 \\
                    \bottomrule
                \end{tabular}

            \end{sc}
        \end{small}
    \end{center}
\end{table}

\subsection{Comparison of kernels}
\label{sec:weights}

A preliminary visual comparison of the convolution kernels is performed. As expected from the evolution of losses, weights and biases are different among the groups of runs, specially for single precision. No relationship or clear tendency could be noticed in terms of which kernels and biases were the most modified and their influences on the different models behavior. Consequently, a global quantitative analysis is proposed and presented further.

For the group of different runs at equal precision, a deviation criterion calculated as the pixel-wise standard deviation normalized by the maximum value among all runs in the group is proposed. For a convolution weight $w$, it is calculated as:

\begin{small}
    \begin{equation}
        \mathrm{deviation}(w) = \frac{
            \mathrm{std}( w^{ \mathrm{run}=i..n } ) }{
                \mathrm{max}( | w^{ \mathrm{run}=i..n } | ) }
        \label{eq:deviation_criterion}
    \end{equation}
\end{small}

\begin{figure}[b!]
    \begin{center}
        \includegraphics[width=7.5cm,trim=0.25cm 0.25cm 0.25cm 0.25cm,clip]
            {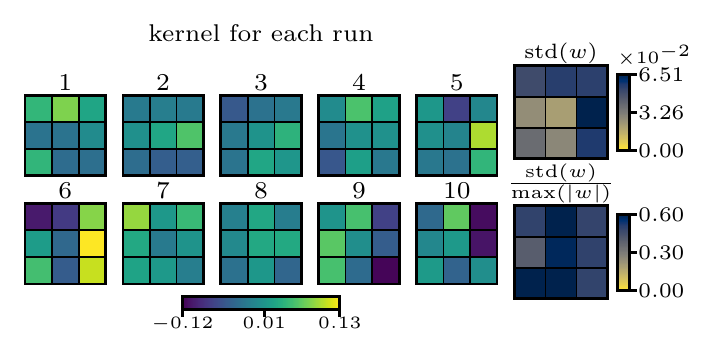}
        \caption{
            Weights, standard deviation and deviation criterion for most modified convolution kernel, runs with single precision.
        }
        \label{fig:kernels_sample_deviation}
    \end{center}
\end{figure}

The calculation is performed for every individual weight, such that the dissimilarity between the models is estimated. An example of the application of the deviation criterion is presented in Figure~\ref{fig:kernels_sample_deviation}, for the kernel that is, on average, the most modified among the single precision runs.

\begin{figure}[t!]
    \begin{center}
        \begin{tabular}{ C{3.75cm} C{3.75cm} }
            {\small {\sc single ($\mathrm{FP32}$) } } & {\small {\sc double ($\mathrm{FP64}$) } } \\[-1ex]
            \includegraphics
                {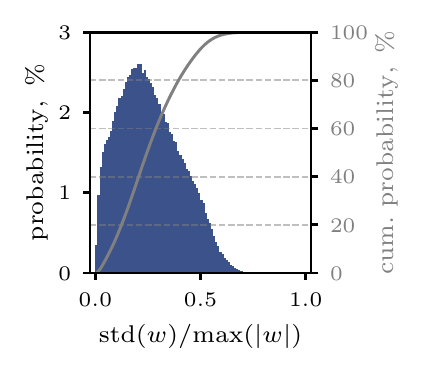}
                &
            \includegraphics
                {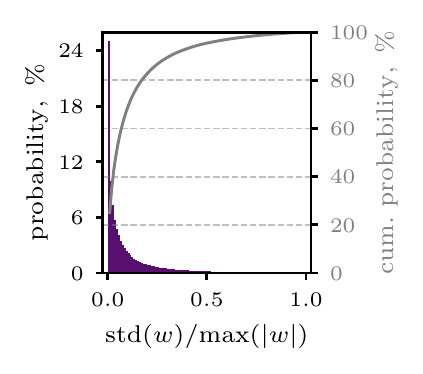}
                \\[-1ex]
        \end{tabular}
        \caption{
            Probability density of the deviation of the convolution weights for the single (left) and double (right) precision runs.}
        \label{fig:graph_weights_deviation_distribution}
    \end{center}
\end{figure}

The distribution of the deviation criteria for all convolution weights is displayed in Figure~\ref{fig:graph_weights_deviation_distribution}, for both the single and double precision. Runs with single precision have a probability mode at 0.21 deviation units, with 80\% of the weights lower than 0.39 units. For the runs with double precision, the mode corresponds to null deviation, with 25.1\% of the weights, and 80\% of the weights are contained within 0.16 of the deviation criterion. These values reinforce both the consequential variability of the single precision runs, probably converged to multiple local optima, and the homogeneity of the trainings with double precision.

\subsection{Featured fields in non-recurrent mode}
\label{sec:featured_fields}

In order to quantify the difference between the models, the featured fields produced for a sample input are also compared. Due to the size of the network, the analysis is limited to the outputs of each scale (quarter, half and full), the latter being the output of the model, used as new input when evaluating the network in recurrent mode (Section~\ref{sec:recurrent}).

\begin{figure}[t!]
    \begin{center}
        \begin{tabular}{ R{0.08cm} C{7.25cm} }
            \rotatebox[origin=c]{90}{\small {\sc single ($\mathrm{FP32}$) } } & 
                \raisebox{-0.5\height}{
                    \includegraphics[trim=0.2cm 0.1cm 0.0cm 0.0cm,clip]
                        {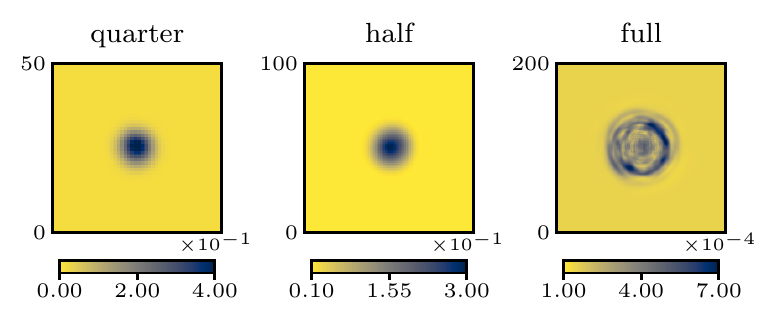}
            } \\
            \rotatebox[origin=c]{90}{\small {\sc double ($\mathrm{FP64}$) } } &
            \raisebox{-0.5\height}{
                \includegraphics[trim=0.2cm 0.2cm 0.0cm 0.55cm,clip]
                    {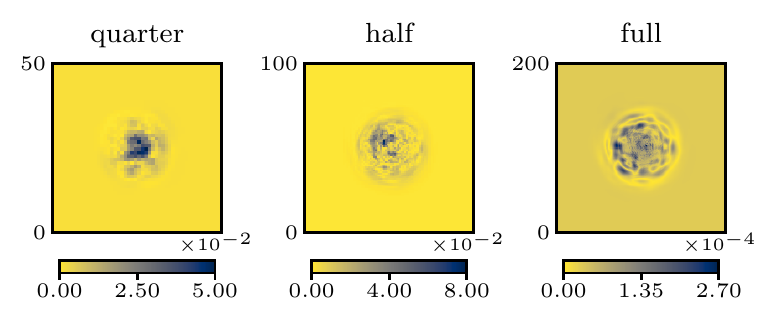}
            } \\
        \end{tabular}

        \caption{
            Pixel-wise deviation of the featured fields (end of each scale) having the start of the Gaussian pulse benchmark as input, for runs with single (top) and double (bottom) precision. Note that the magnitudes do not correspond to the any physical quantity and colormap ranges are different for each field.
        }
        \label{fig:fields_featured_fields_deviation}
    \end{center}
\end{figure}

The first 4 frames (simulation timestesps 0, 4, 8 and 12) of the Gaussian pulse benchmark are selected as input. Quantification of the deviation among different models is made using the criterion defined on Equation~\ref{eq:deviation_criterion}, here applied to each value that composes the fields. Analysis is split between the runs in single and double precision and the results are shown in Figure~\ref{fig:fields_featured_fields_deviation}.

Even if restricted by the small size of the dataset, this analysis leads to two conclusions. First, a large variation (order of 10\% of the maximum value) is obtained for both the quarter and half scales, indicating a rather high variability of the response for the hidden layers. Due to use of non-deterministic  algorithms and the stochastic aspect of the optimization, the fact that each run leads to a unique result is expected. Nevertheless, it is remarkable that every run leads to a completely different internal dynamics, being the features observed at hidden layers different in both shape and amplitude, what is represented by the uniform patches of high deviation. Second, it is clear that double precision runs are more uniform, with deviation values lower by around one decade, indicating that the global behavior of the network is reproduced by all models despite not using deterministic algorithms.

\section{Performance of recurrent estimation}
\label{sec:recurrent}

Several models have been obtained from the non-deterministic training, leading to small differences of the output in single-prediction mode. This section investigates how the non-determinism inherent to the model affects the long-time prediction in recurrent mode. This is evaluated by calculating the total loss considering the simulated and estimated fields, stating from the single-prediction mode (recurrence number ${r = 0}$) and reuse the output as new input (recurrent mode, ${r > 0}$). Note that the recurrent fields are not used as targets when training the neural network, so the use of the name ``loss'' is an extrapolation of its concept.

Since the loss is calculated considering normalized fields, it is not a direct quantification of the estimation accuracy. For that, the root-mean-square error ($\mathrm{RMSE}$), issued from physical units, is also calculated:

\begin{small}
    \begin{equation}
        \mathrm{RMSE}(r) = \sqrt{
            {\textstyle \sum_{k=1}^M} \left[ \rho\rq_k(r) - \tilde{\rho}\rq_k(r) \right]^2 / M 
            }
        \label{eq:RMSE}
    \end{equation}
\end{small}

For convenience, the $\mathrm{RMSE}$ values are normalized by the pulse amplitude $\varepsilon$. Recursive analysis is performed for the three benchmarks (Section~\ref{sec:test_benchmarks}) and the random pulse database (Section~\ref{sec:test_random}).

\subsection{Benchmarks}
\label{sec:test_benchmarks}

\begin{figure}[b!]
    \begin{center}
        \begin{tabular}{ l }
            \centerline{ {\small {\sc single precision ($\mathrm{FP32}$) } } } \\[-1ex] 
            \includegraphics[trim=0.05cm 0.80cm 0 0,clip]
                {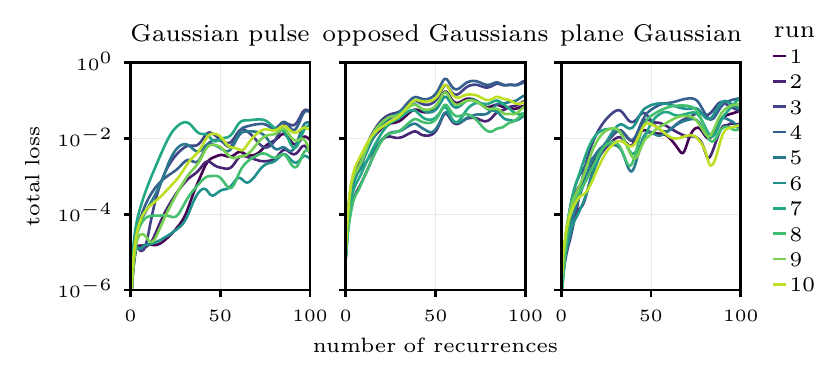}
                \\[-0.5ex]
            \includegraphics[trim=0 0.5cm 0 0.5cm,clip]
                {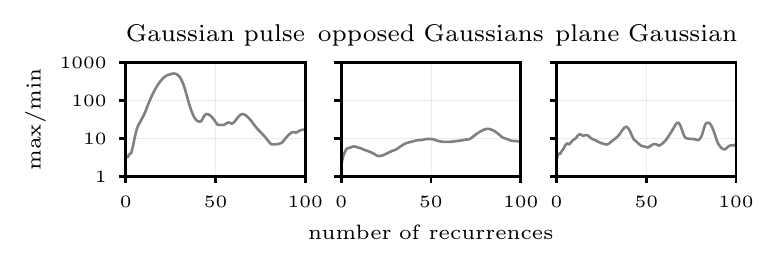}
                \\

            \centerline{ {\small {\sc double precision ($\mathrm{FP64}$) } } } \\
            \includegraphics[trim=0.05cm 0.80cm 0 0.5cm,clip]
                {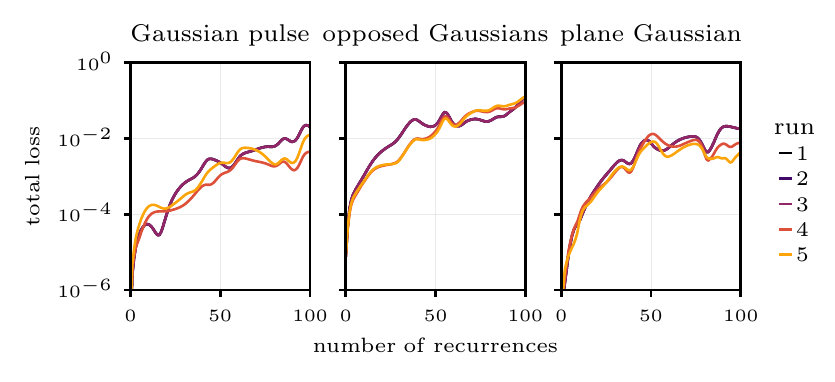}
                \\[-0.5ex]
            \includegraphics[trim=0 0 0 0.5cm,clip]
                {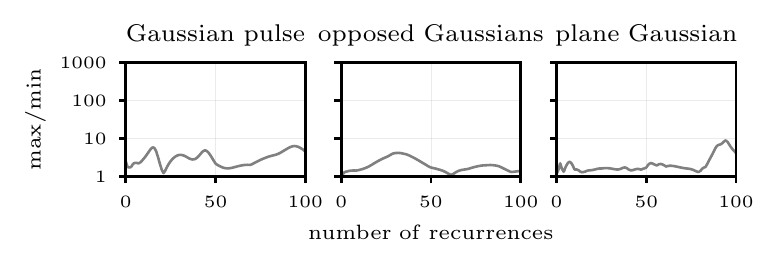}
                \\
        \end{tabular}
        \caption{
            Evolution of total loss and extrema ratio for the recurrent test of benchmarks for models trained with single (top) and double (bottom) precision; curves for runs 1 to 3 with double precision are superposed.
        }
        \label{fig:graph_recurrent_error_benchmarks}
    \end{center}
\end{figure}

Evolution of the total loss in the recurrent estimation of the benchmarks for the models trained with single and double precision are presented in Figure~\ref{fig:graph_recurrent_error_benchmarks}. The capacity of the neural network to capture the physical phenomenon is evidenced by the fact that similar performance is obtained for features that were not present in the training database (negative and planar pulses).

\begin{figure}[t!]
    \begin{center}
        \includegraphics[trim=0.2cm 0.2cm 0 0.1cm,clip]
            {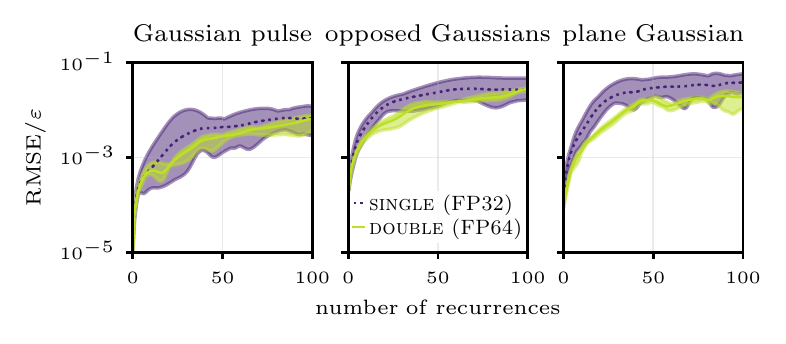}
        \caption{
            Evolution of the root-mean-square error normalized by the pulse amplitude for the recurrent test of benchmarks for models trained with single (dotted line) and double (full line) precision; central line represents the average among the runs and band indicates the minimum to maximum range.
        }
        \label{fig:graph_recurrent_error_singleAndDouble}
    \end{center}
\end{figure}

Loss (and consequentially the error) remains relatively small even after 100 recurrences, showing the overall quality of the estimation. The shape of the curves are similar when comparing the runs with same floating-point precision or the two groups of runs. Losses approach ${\mathcal{L} = 0.01}$ in the two graphs, that is, both single and double precision resulted in similar overall preciseness for the tested benchmarks. The quality and similarity of the two groups of runs is also present when comparing the evolution of the error criterion $\mathrm{RMSE}$, presented in Figure~\ref{fig:graph_recurrent_error_singleAndDouble}.

In spite of the general quality and similar behavior, an important range exists among the different runs for the single precision trainings. Losses’ maximum to minimum ratio (Figure~\ref{fig:graph_recurrent_error_benchmarks}) peaked at 550 for single precision trainings, and curves fluctuate around 10. For double precision, such ratio was always lower than 10 for all the 3 benchmarks, with an average value of 3.2 for the Gaussian pulse.

In conclusion, both single and double precision models yield a similar behavior on the 3 benchmarks, with a slightly better accuracy when using double precision. However, the comparison of several runs explicit that a high variability is present, specially for models obtained with single precision. Results also highlight that the recursive mode amplifies differences due to the non-deterministic training, which makes reproducibility a crucial element to take into account when applying neural network to recursive tasks.

\subsection{Random pulses}
\label{sec:test_random}

\begin{figure}[b!]
    \begin{center}
        \begin{tabular}{ C{0.8cm} C{4.05cm} C{2.30cm} } 
            & {\small {\sc single ($\mathrm{FP32}$) } } & {\small {\sc double ($\mathrm{FP64}$) } } \\
        \end{tabular}
        \vskip -0.75cm
        \begin{tabular}{ R{5.15cm} L{2.4cm} }
            \includegraphics[trim=0.2cm 0.0cm 0 0.0cm,clip]
                {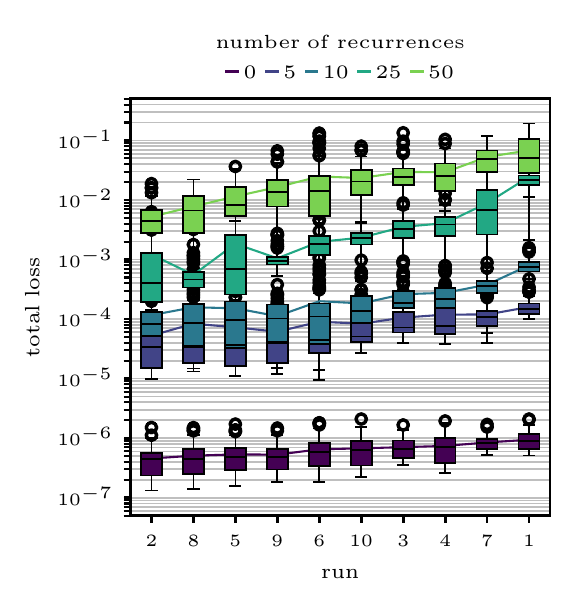}
            &
            \includegraphics[trim=1.15cm 0.0cm 0 0.0cm,clip]
                {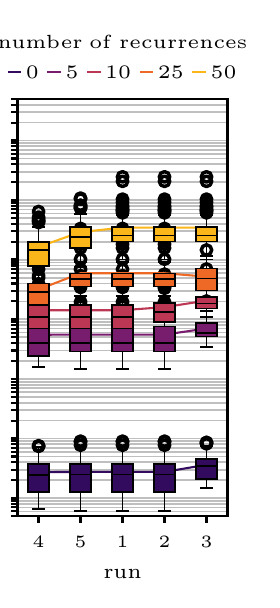}
            \\
        \end{tabular}
        \caption{
            Box plots of the total loss for the different models obtained with single precision (left) and double precision (right), considering the random pulse database at multiple number of recurrences. The central lines indicates the median value, the box limits represent first and third quartiles, the whiskers represents the median $\pm$ 1.5 times the interquartile range and circles represent the outliers; continuous lines connects the average for each run.
        }
        \label{fig:graph_recurrent_error_random_boxplots}
    \end{center}
\end{figure}

A statistical approach is used to compare the recurrent test results for the random database. For each simulation, the same recursive test used for the benchmarks database is performed. The evolution of the testing losses are represented for a subset of recurrences (0, 5, 10, 25, 50) in Figure~\ref{fig:graph_recurrent_error_random_boxplots}. Each boxplot describes the distribution of the loss for the 100 simulations that compose the database, at the given number of recurrences. Note that runs 1 to 3 with double precision do not return identical distributions despite having the same loss (Figure~\ref{fig:graph_loss_convergence}).

For all models, the median loss lies lower than the converging amplitude in training ($\mathcal{L}$ around ${2 \times 10^{-6}}$). At the recurrence 0 (evaluation of the model as it was trained), comparison of average losses returns a factor 2.06 between the worst (run 1) and the best (run 2) models trained with single precision and 1.33 for double precision (runs 3 and 4, respectively). At a first look, this variation seems to be a direct outcome of the losses’ ratios for the different runs (Tables~\ref{tab:models_single} and~\ref{tab:models_double}). However, the current ratios do not come from the same models that delimit minimum and maximum losses in learning phase. Considering validation losses, worst and best models are runs 9 and 6 for single precision and runs 5 and 4 for double precision.

\begin{figure}[t!]
    \begin{center}
        \includegraphics[trim=0.0cm 0.0cm 0.0cm 0.0cm,clip]
            {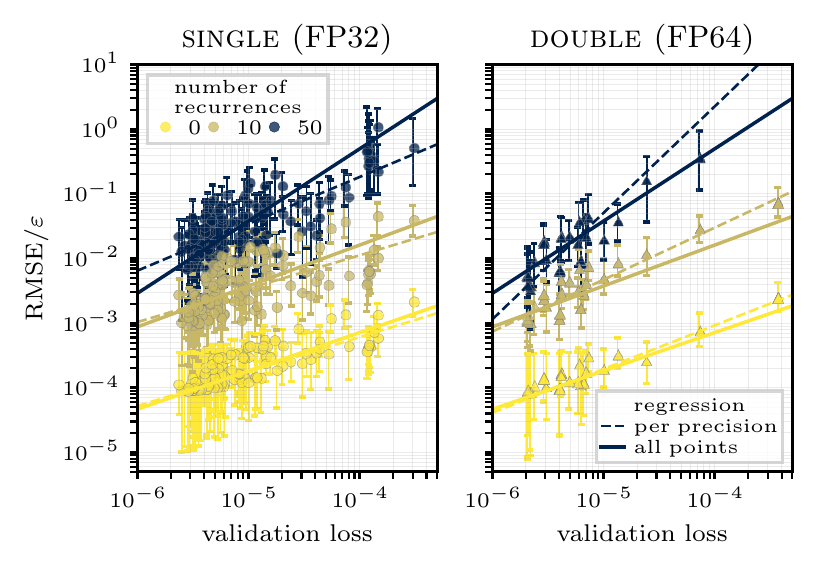}
        \caption{
            Average recurrent test error versus validation loss for all simulations in random pulse database at multiple numbers of recurrence, single precision (left) and double precision (right) trainings, for models obtained at every 125 epochs; error bars indicate minimum to maximum values.
        }
        \label{fig:graph_validation_loss_vs_test_error_mean}
    \end{center}
\end{figure}

The direct relationship between the models’ validation losses and their performance on recurrent mode is examined. Scope of analysis is extended by considering the complete set of models available per run (one every 125 epochs, 12 models per run) and not only the best states as previously. Once the focus is on the accuracy of the estimation, $\mathrm{RMSE}$ is considered instead of the loss. In Figure~\ref{fig:graph_validation_loss_vs_test_error_mean}, average error and min-max bands for the 100 simulations in the database are plotted against validation losses for recurrence numbers 0, 10 and 50. Linear regression coefficients for the single and double precision models and the combined set of points are presented in Table~\ref{tab:error_regression}.

\begin{table}[b!]
    \caption{Linear regression of testing error ($\mathrm{RMSE}$) as a function of the model's validation loss ($\mathcal{L}^\mathrm{val}$) for multiple recurrences numbers, $\mathrm{R}^2$ is the determination coefficient. Wald Test with {$t$-distribution} returned two-sided null $p$-values ($\mathrm{H_0}$:~null~slope) for all regressions.}
    \label{tab:error_regression}
    \begin{center}
        \begin{small}
            \begin{sc}
                \begin{tabular}{ C{1.2cm}ccccc }
                    \toprule
                    \multicolumn{5}{c}{
                        $\mathrm{RMSE(r)}/\varepsilon = 
                            (\mathcal{L}^\mathrm{val})^A \times B$ } \\
                    \midrule    
                    & $r$            & 0 & 10  & 50 \\
                    \midrule
                    \multirow{3}{*}{ \shortstack{ \small {\sc single } \\ ($\mathrm{FP32}$) } }
                    & $A$            & 0.533 & 0.517 & 0.725 \\
                    & $B$            & 8.22${\scriptstyle \times 10^{-2}}$ & 1.31${\scriptstyle \times 10^{0}}$ & 1.45${\scriptstyle \times 10^{2}}$ \\
                    \cmidrule{2-5}
                    & $\mathrm{R}^2$ & 0.36 & 0.44 & 0.55 \\
                    \midrule
                    \multirow{3}{*}{ \shortstack{ \small {\sc double } \\ ($\mathrm{FP64}$) } }
                    & $A$            & 0.676 & 0.804 & 1.647 \\
                    & $B$            & 4.67${\scriptstyle \times 10^{-1}}$ & 4.91${\scriptstyle \times 10^{1}}$ & 8.71${\scriptstyle \times 10^{6}}$ \\
                    \cmidrule{2-5}
                    & $\mathrm{R}^2$ & 0.69 & 0.84 & 0.90 \\
                    \midrule
                    \multirow{3}{*}{ \shortstack{ \small {\sc all } \\ {\sc points } } }
                    & $A$            & 0.593 & 0.633 & 1.118 \\
                    & $B$            & 1.68${\scriptstyle \times 10^{-1}}$ & 5.50${\scriptstyle \times 10^{0}}$ & 1.47${\scriptstyle \times 10^{4}}$ \\
                    \cmidrule{2-5}
                    & $\mathrm{R}^2$ & 0.54 & 0.53 & 0.68 \\
                    \bottomrule
                \end{tabular}

            \end{sc}
        \end{small}
    \end{center}
\end{table}

Evolution of the coefficients $A$ and $B$ with the recurrence number $r$ stresses the increase of the error associated with the recurrent evaluation. After several recurrences, small deviations observed at the time of training, here quantified by the model’s training loss, are significantly amplified. Despite the global trend, variability is elevated for any given abscissa and such conclusion can not be used to directly compare models with close validation loss as observed in the analysis of the box-plots.

The regression coefficients indicate that models trained with double precision are, on average, more prone to amplify the initial errors, suggesting that the accumulation of error is larger for double precision. Such regression curves are issued from a smaller and much less spread dataset (points concentrated at $\mathcal{L}^\mathrm{val} < 10^{-5}$, many of them superposed), thus, less significant and more subjected to the influence of outliers. Besides, recurrent estimation (until ${r = 100}$) of the benchmarks performed with single precision for a model trained with double precision (run 1) returned a disparity in $\mathrm{RMSE}$ lower than 1\% when compared to the double precision inference. In other words, there is no apparent increase in error accumulation due to the use of single or double precision in testing. This indicates that inference may be performed with lower precision with reduced impact in accuracy while gaining in performance (factor 1/3 in calculation time for current application).

The test with a broader number of simulations showed that there is an important range of variation associated with the different models. Although the overall quality of the approximation is almost unmodified when training with single or double precision, the latter produces models that return more uniform results when tested, what is directly associated with its similarity at null recurrence.

\section{Final discussion and conclusions}
\label{sec:conclusions}

Reproducibility of a fully convolutional neural network used to model a time-evolving physical system has been assessed by performing the same training several times under GPU non-determinism. Use of deterministic calculations is, up-to-now, not always available and may also not be consistent among distinct hardware. Since the training non-determinism is associated with the truncation error, tests are performed with single (32-bits) and double (64-bits) precision.

Runs with single precision revealed a large variation in terms of parameters and results. For the current test cases, the use of double precision was capable of limiting the variability of kernels by a factor 2, and constrained the error in the recurrent analysis, while increasing the models accuracy slightly. Compared to the single precision runs, computation cost was multiplied by two in terms of memory and by three for the training time. This important gain in cost may be prohibitive for more complex networks and equations, such as in the evaluation of three-dimensional problems, and a hybrid strategy - multi or mixed-precision~\cite{Micikevicius2018} - could also be envisaged. Also, it was observed that performing training with double and  inference with single precision resulted in negligible loss of  accuracy.

Results presented here are for a single neural network architecture and PDE, a natural extension of this work would be performing similar analysis for different architectures and equations. Candidate strategies to the increase of the precision that could also reduce the variability, but were not tested in current work, are: to impose hidden layer reproducibility by adding intermediary fields error terms in loss, what could depreciate the estimation quality; the use of physics informed network~\cite{Meng2020a}; or the use of ``long-term loss''~\cite{Tompson2017}. Also, the analysis can be extended by evaluating whether the variability and the influence of the floating-point precision is similar with a truly recurrent neural network, such as LSTM.

The performed analysis has shown a high variability associated to a typical fully CNN model, trained to reproduce time evolving fields. Based on the highlighted behavior, it is recommended that analogous work should be performed whenever possible when modeling physical systems, where small deviations may lead to divergences or be the onset of instabilities. Reproducibility must be  continuously on focus, particularly if one aims at producing robust, application oriented models such as solver accelerators.

\section*{Acknowledgments}
\label{sec:acknowledgments}

Authors gratefully acknowledge the financing by the French Government Defense procurement and technology agency DGA (\textbf{D}irectorate \textbf{G}eneral of \textbf{A}rmaments) in the project POLA3.

\bibliographystyle{plain}
\bibliography{paper_CNN_reproducibility_2021}

\begin{thebibliography}{10}

\bibitem{Alguacil2020}
Antonio Alguacil, Michaël Bauerheim, Marc~C. Jacob, and Stephane Moreau.
\newblock {\em Predicting the Propagation of Acoustic Waves using Deep
  Convolutional Neural Networks}.
\newblock 2020.

\bibitem{Berg2018}
Jens Berg and Kaj Nyström.
\newblock A unified deep artificial neural network approach to partial
  differential equations in complex geometries.
\newblock {\em Neurocomputing}, 317:28--41, nov 2018.

\bibitem{Brunton2020}
Steven~L. Brunton, Bernd~R. Noack, and Petros Koumoutsakos.
\newblock Machine learning for fluid mechanics.
\newblock {\em Annual Review of Fluid Mechanics}, 52(1):477--508, 2020.

\bibitem{Chou2020}
Yuan~Hsi Chou, Christopher Ng, Shaylin Cattell, Jeremy Intan, Matthew~D.
  Sinclair, Joseph Devietti, Timothy~G. Rogers, and Tor~M. Aamodt.
\newblock Deterministic atomic buffering.
\newblock In {\em 2020 53rd Annual {IEEE}/{ACM} International Symposium on
  Microarchitecture ({MICRO})}. {IEEE}, oct 2020.

\bibitem{Denton2015}
Emily~L Denton, Soumith Chintala, Arthur Szlam, and Rob Fergus.
\newblock {D}eep {G}enerative {I}mage {M}odels using a {L}aplacian {P}yramid of
  {A}dversarial {N}etworks.
\newblock In C.~Cortes, N.~Lawrence, D.~Lee, M.~Sugiyama, and R.~Garnett,
  editors, {\em Advances in Neural Information Processing Systems}, volume~28,
  pages 1486--1494. Curran Associates, Inc., 2015.

\bibitem{Fotiadis2020}
Stathi Fotiadis, Eduardo Pignatelli, Mario~Lino Valencia, Chris Cantwell, Amos
  Storkey, and Anil~A. Bharath.
\newblock Comparing recurrent and convolutional neural networks for predicting
  wave propagation.
\newblock In {\em {ICLR 2020 Workshop on Integration of Deep Neural Models and
  Differential Equations}}, 2020.

\bibitem{He2015}
Kaiming He, Xiangyu Zhang, Shaoqing Ren, and Jian Sun.
\newblock Delving deep into rectifiers: Surpassing human-level performance on
  imagenet classification.
\newblock In {\em Proceedings of the IEEE International Conference on Computer
  Vision (ICCV)}, December 2015.

\bibitem{Iakymchuk2015}
Roman Iakymchuk, David Defour, Sylvain Collange, and Stef Graillat.
\newblock Reproducible and accurate matrix multiplication.
\newblock In {\em International Symposium on Scientific Computing, Computer
  Arithmetic, and Validated Numerics}, pages 126--137. Springer, 2015.

\bibitem{Ivie2018}
Peter Ivie and Douglas Thain.
\newblock {Reproducibility in Scientific Computing}.
\newblock {\em {{ACM} Computing Surveys}}, 51(3):1--36, jul 2018.

\bibitem{Jezequel2015}
Fabienne J{\'{e}}z{\'{e}}quel, Jean-Luc Lamotte, and Issam Saïd.
\newblock Estimation of numerical reproducibility on {CPU} and {GPU}.
\newblock In {\em Proceedings of the 2015 Federated Conference on Computer
  Science and Information Systems}. {IEEE}, oct 2015.

\bibitem{Jorda2019}
Marc Jorda, Pedro Valero-Lara, and Antonio~J. Pena.
\newblock Performance evaluation of {cuDNN} convolution algorithms on {NVIDIA}
  volta {GPUs}.
\newblock {\em {IEEE} Access}, 7:70461--70473, 2019.

\bibitem{Kingma2015}
D.P. Kingma and L.J. Ba.
\newblock Adam: {A} {M}ethod for {S}tochastic {O}ptimization.
\newblock May 2015.
\newblock {I}nternational {C}onference on {L}earning {R}epresentations
  ({ICLR}).

\bibitem{Kruger2016}
T.~Kr{\"u}ger, H.~Kusumaatmaja, A.~Kuzmin, O.~Shardt, G.~Silva, and E.M.
  Viggen.
\newblock {\em {T}he {L}attice {B}oltzmann {M}ethod: {P}rinciples and
  {P}ractice}.
\newblock Graduate Texts in Physics. Springer International Publishing, 2016.

\bibitem{Palabos2020}
Jonas Latt, Orestis Malaspinas, Dimitrios Kontaxakis, Andrea Parmigiani, Daniel
  Lagrava, Federico Brogi, Mohamed~Ben Belgacem, Yann Thorimbert, Sébastien
  Leclaire, Sha Li, Francesco Marson, Jonathan Lemus, Christos Kotsalos,
  Raphaël Conradin, Christophe Coreixas, Rémy Petkantchin, Franck Raynaud,
  Joël Beny, and Bastien Chopard.
\newblock {P}alabos: {P}arallel {L}attice {B}oltzmann {S}olver.
\newblock {\em Computers \& Mathematics with Applications}, 2020.

\bibitem{Lee2019}
Sangseung Lee and Donghyun You.
\newblock Data-driven prediction of unsteady flow over a circular cylinder
  using deep learning.
\newblock {\em Journal of Fluid Mechanics}, 879:217--254, sep 2019.

\bibitem{Lee2019a}
Sangseung Lee and Donghyun You.
\newblock Mechanisms of a convolutional neural network for learning
  three-dimensional unsteady wake flow.
\newblock In {\em 72nd Annual Meeting of the APS Division of Fluid Dynamics}.
  {American Physical Society}, nov 2019.

\bibitem{Li2020}
Liam Li and Ameet Talwalkar.
\newblock {Random Search and Reproducibility for Neural Architecture Search}.
\newblock In Ryan~P. Adams and Vibhav Gogate, editors, {\em {Proceedings of The
  35th Uncertainty in Artificial Intelligence Conference}}, volume 115 of {\em
  {Proceedings of Machine Learning Research}}, pages 367--377, Tel Aviv,
  Israel, 22--25 Jul 2020. {PMLR}.

\bibitem{Mathieu2016}
Micha{\"{e}}l Mathieu, Camille Couprie, and Yann {LeCun}.
\newblock {D}eep multi-scale video prediction beyond mean square error.
\newblock In {Y}oshua {B}engio and {Y}ann {LeCun}, editors, {\em 4th
  {I}nternational {C}onference on {L}earning {R}epresentations, {ICLR} 2016,
  {S}an {J}uan, {P}uerto {R}ico, {M}ay 2-4, 2016, {C}onference {T}rack
  {P}roceedings}, 2016.

\bibitem{Meng2020a}
Xuhui Meng, Zhen Li, Dongkun Zhang, and George~Em Karniadakis.
\newblock {PPINN}: Parareal physics-informed neural network for time-dependent
  {PDEs}.
\newblock {\em Computer Methods in Applied Mechanics and Engineering},
  370:113250, oct 2020.

\bibitem{Micikevicius2018}
Paulius Micikevicius, Sharan Narang, Jonah Alben, Gregory Diamos, Erich Elsen,
  David Garcia, Boris Ginsburg, Michael Houston, Oleksii Kuchaiev, Ganesh
  Venkatesh, and Hao Wu.
\newblock Mixed precision training.
\newblock In {\em International Conference on Learning Representations}, 2018.

\bibitem{Nagarajan2018}
Prabhat Nagarajan, Garrett Warnell, and Peter Stone.
\newblock The impact of nondeterminism on reproducibility in deep reinforcement
  learning.
\newblock In {\em {2nd Reproducibility in Machine Learning Workshop at ICML
  2018}}, 2018.

\bibitem{cuda_CUDNN_doc}
{NVIDIA Corporation}.
\newblock {CUDA} {T}oolkit {D}ocumentation v11.1.0: 2.1.4. {R}esults
  reproducibility.
\newblock
  \url{https://docs.nvidia.com/cuda/archive/11.1.0/cublas/index.html#cublasApi_reproducibility},
  2020.
\newblock [Online; accessed 18-January-2021].

\bibitem{cuda_cuBLAS_doc}
{NVIDIA Corporation}.
\newblock cu{DNN} {D}eveloper {G}uide: Chapter 8. {R}eproducibility
  (determinism).
\newblock
  \url{https://docs.nvidia.com/deeplearning/cudnn/pdf/cuDNN-Developer-Guide.pdf},
  2020.
\newblock [Online; accessed 18-January-2021].

\bibitem{pyTorch}
Adam Paszke, Sam Gross, Francisco Massa, Adam Lerer, James Bradbury, Gregory
  Chanan, Trevor Killeen, Zeming Lin, Natalia Gimelshein, Luca Antiga, Alban
  Desmaison, Andreas Kopf, Edward Yang, Zachary DeVito, Martin Raison, Alykhan
  Tejani, Sasank Chilamkurthy, Benoit Steiner, Lu~Fang, Junjie Bai, and Soumith
  Chintala.
\newblock {P}y{T}orch: {A}n {I}mperative {S}tyle, {H}igh-{P}erformance {D}eep
  {L}earning {L}ibrary.
\newblock In H.~Wallach, H.~Larochelle, A.~Beygelzimer, F.~d\textquotesingle
  Alch\'{e}-Buc, E.~Fox, and R.~Garnett, editors, {\em {A}dvances in {N}eural
  {I}nformation {P}rocessing {S}ystems 32}, pages 8024--8035. {C}urran
  {A}ssociates, Inc., 2019.

\bibitem{Peng2011}
R.~D. Peng.
\newblock Reproducible research in computational science.
\newblock {\em Science}, 334(6060):1226--1227, dec 2011.

\bibitem{Pineau2020}
Joelle Pineau, Philippe Vincent-Lamarre, Koustuv Sinha, Vincent Larivière,
  Alina Beygelzimer, Florence d'Alché Buc, Emily~B. Fox, and Hugo Larochelle.
\newblock Improving reproducibility in machine learning research (a report from
  the neurips 2019 reproducibility program).
\newblock {\em CoRR}, abs/2003.12206, 2020.

\bibitem{Renard2020}
F{\'{e}}lix Renard, Soulaimane Guedria, Noel~De Palma, and Nicolas Vuillerme.
\newblock Variability and reproducibility in deep learning for medical image
  segmentation.
\newblock {\em {Scientific Reports}}, 10(1), aug 2020.

\bibitem{Ronneberger2015}
Olaf Ronneberger, Philipp Fischer, and Thomas Brox.
\newblock U-net: Convolutional networks for biomedical image segmentation.
\newblock In {\em International Conference on Medical image computing and
  computer-assisted intervention}, pages 234--241. Springer, 2015.

\bibitem{Seznec2018}
Mickael Seznec, Nicolas Gac, Andre Ferrari, and Francois Orieux.
\newblock A study on convolution using half-precision floating-point numbers on
  {GPU} for radio astronomy deconvolution.
\newblock In {\em 2018 {IEEE} International Workshop on Signal Processing
  Systems ({SiPS})}. {IEEE}, oct 2018.

\bibitem{Sirignano2018}
Justin Sirignano and Konstantinos Spiliopoulos.
\newblock {DGM}: {A} deep learning algorithm for solving partial differential
  equations.
\newblock {\em {Journal of Computational Physics}}, 375:1339 -- 1364, 2018.

\bibitem{Sorteberg2018}
Wilhelm Sorteberg, Stef Garasto, Alison Pouplin, Chris Cantwell, and
  Anil~Anthony Bharath.
\newblock Approximating the solution to wave propagation using deep neural
  networks.
\newblock In {\em Proceedings of the 32nd Conference on Neural Information
  Processing Systems (NIPS 2018)}, Montreal, Canada, Dec 7 2018.

\bibitem{Tompson2017}
Jonathan Tompson, Kristofer Schlachter, Pablo Sprechmann, and Ken Perlin.
\newblock Accelerating {E}ulerian fluid simulation with convolutional networks.
\newblock In Doina Precup and Yee~Whye Teh, editors, {\em Proceedings of the
  34th International Conference on Machine Learning}, volume~70 of {\em
  Proceedings of Machine Learning Research}, pages 3424--3433, International
  Convention Centre, Sydney, Australia, 06--11 Aug 2017. PMLR.

\bibitem{pyTorch_determinisitic_doc}
{Torch contributors}.
\newblock {P}y{T}orch documentation.
\newblock
  \url{https://pytorch.org/docs/stable/generated/torch.set_deterministic.html},
  2019.
\newblock [Online; accessed 18-January-2021].

\bibitem{Whitehead2011}
Nathan Whitehead and Alex Fit-Florea.
\newblock Precision \& performance: {F}loating point and {IEEE} 754 compliance
  for {NVIDIA} {GPUs}.
\newblock Technical report, NVIDIA Corporation, 2011.

\end{thebibliography}

\end{document}